\documentclass{article}

 \usepackage[dblblindworkshop, final]{neurips_2025}
\workshoptitle{LAW 2025: Bridging Language, Agent, and World Models for Reasoning and Planning}

\usepackage[utf8]{inputenc} 
\usepackage[T1]{fontenc}    
\usepackage{booktabs}       
\usepackage{amsfonts}       
\usepackage{nicefrac}       
\usepackage{microtype}      
\usepackage{xcolor}         

\usepackage{tabularx}
\usepackage{makecell}
\usepackage{placeins}

\usepackage{graphicx} 
\usepackage{amsmath} 
\usepackage[hyphens]{url} 
\usepackage{caption} 
\usepackage{subcaption} 
\usepackage{natbib} 
\usepackage{enumitem}
\usepackage{amssymb}
\setlist[itemize]{left=1.5em}
\setlist[enumerate]{left=1.5em}
\usepackage{float}
\usepackage[colorlinks=true, linkcolor=blue, citecolor=blue]{hyperref}
\usepackage{adjustbox}

\usepackage{ragged2e} 
\usepackage{longtable} 
\usepackage{array} 
\usepackage{geometry}
\usepackage{tikz}
\usepackage{multirow}

\newcolumntype{R}{>{\RaggedRight\arraybackslash}X}
\title{Agentic Design Patterns: A System-Theoretic Framework}

\author{%
  Minh-Dung Dao \thanks{Equal contribution.}~~\footnotemark[2]\\
  University College Cork \\
  College Rd, Cork, Ireland T12 K8AF \\
  \texttt{123122658@umail.ucc.ie} \\
  \And
  Quy Minh Le \footnotemark[1]~~\footnotemark[2] \\
  Vietnam National University \\
  Xuan Thuy St, Hanoi, Vietnam 10000 \\
  \texttt{22028190@vnu.edu.vn} \\
  \AND
  Hoang Thanh Lam \\
  IBM Research Ireland \\
  182 Pearse Street, Dublin 2, Ireland D02 F6N2 \\
  \And
  Duc-Trong Le \\
  Vietnam National University \\
  Xuan Thuy St, Hanoi, Vietnam 10000 \\
  \AND
  Quoc-Viet Pham \\   
  Trinity College Dublin \\
  Dublin 2, Ireland D02 W272 \\
  \And
  Barry O'Sullivan \\
  University College Cork \\
  College Rd, Cork, Ireland T12 K8AF \\
  \AND
  Hoang D. Nguyen \thanks{Corresponding authors.} \\
  University College Cork \\
  College Rd, Cork, Ireland T12 K8AF \\
  \texttt{hn@cs.ucc.ie}
}

\begin{document}

\maketitle

\begin{abstract}
  With the development of foundation model (FM), agentic AI systems are getting more attention, yet their inherent issues like hallucination and poor reasoning, coupled with the frequent ad-hoc nature of system design, lead to unreliable and brittle applications. Existing efforts to characterise agentic design patterns often lack a rigorous systems-theoretic foundation, resulting in high-level or convenience-based taxonomies that are difficult to implement. This paper addresses this gap by introducing a principled methodology for engineering robust AI agents. We propose two primary contributions: first, a novel system-theoretic framework that deconstructs an agentic AI system into five core, interacting functional subsystems: Reasoning \& World Model, Perception \& Grounding, Action Execution, Learning \& Adaptation, and Inter-Agent Communication. Second, derived from this architecture and directly mapped to a comprehensive taxonomy of agentic challenges, we present a collection of 12 agentic design patterns. These patterns — categorised as Foundational, Cognitive \& Decisional, Execution \& Interaction, and Adaptive \& Learning — offer reusable, structural solutions to recurring problems in agent design. The utility of the framework is demonstrated by a case study on the ReAct framework, showing how the proposed patterns can rectify systemic architectural deficiencies. This work provides a foundational language and a structured methodology to standardise agentic design communication among researchers and engineers, leading to more modular, understandable, and reliable autonomous systems.
\end{abstract}

\section{Introduction}


Foundation model (FM) creates a revolution in Artificial Intelligence (AI); AI systems can demonstrate behaviours reminiscent of natural entities with cognitive skills, such as remembering, reasoning, thinking and writing creatively \cite{naveed2025comprehensive}. They have enabled a wide variety of applications in different fields and changed the paradigm of research on intelligent systems. However, FMs face various problems that hinder their capabilities and usefulness for practical applications, such as catastrophic forgetting, hallucination, bias, and incapacity of slow thinking \cite{kaddour2023challenges}. Recently, there has been a growing interest from both industry and academia on agentic AI systems \cite{acharya2025agentic}, with FMs at their cores and equipped with external tools (e.g., web searching and code execution) and abilities (e.g., memorising and planning). This approach allows systems to tackle difficult problems, interact with external environments, and make decisions with a certain level of autonomy. Moreover, a combination of many intelligent agents that interact with each other following certain structures, strategies, and coordination protocols creates a multi-agent system (MAS), in which agents can communicate, share information, orchestrate, and act toward a set of collective goals \cite{tran2025multi}. These ideas about agentic systems and MAS originate from individual and collective human intelligence in society, enabled with equipments and collaborative mechanisms to perform different activities from daily routines to complex scientific thinking.

There have been several different attempts to formulate an encompassing definition of agentic AI systems, as well as establish common strategies to deal with the inherent aforementioned problems of the core FMs and problems arising from additional tools, capabilities, and interactions. Such strategies are often known as agentic design patterns (ADPs), and there have been different efforts to organise ADPs into structures \cite{Ng2024Agentic, liu2025agent}. However, these attempts lack a systems-theoretic basis to facilitate a rigorous understanding of agentic AI and/or are mostly convenience-based taxonomies that originate mainly from observations of practical applications. Furthermore, the proposed ADPs are often high-level and their organisation is complicated, making them less useful for direct implementation. There is also little to no connection from existing ADPs to well-established software design patterns that are widely implemented in software systems \cite{gamma1995design}. A systematic design approach is necessary to understand the purpose of different components, as well as create a collection of design patterns that allow solving different classes of problems and be able to apply straightforwardly in creating new agentic AI system or improving existing ones \cite{miehling2025agenticaineedssystems}.   


This paper introduces a principled engineering discipline for agentic AI systems to address the brittleness of current ad-hoc approaches. To achieve this, we embark on a structured inquiry to answer two fundamental research questions:
\begin{enumerate}
    \item How can we formulate agentic AI with a rigorous, systems-theoretic foundation that moves beyond monolithic FM-centric designs?
    \item What are the systemic classes of problems that undermine agent reliability, and what specific, reusable design patterns can provide structural solutions?
\end{enumerate}


To answer these questions, the paper is organised as follows. We first establish the problem domain by reviewing foundational concepts and identifying the critical gap in current methodologies. We then systematically categorise the challenges plaguing FM-based agents into five classes, from World Modelling to Collaboration Mechanisms, providing a clear problem map.

In response to our first research question, we introduce our core contribution: a novel system-theoretic framework that conceptualises an agent as a layered organisation of five primary functional subsystems: Reasoning \& World Model, Perception \& Grounding, Action Execution, Learning \& Adaptation, and Inter-Agent Communication. This architecture provides the theoretical foundation for the construction and analysis of agentic systems in a principled manner.

Addressing our second research question, we derive from this framework a comprehensive collection of 12 agentic design patterns. Each pattern, such as Intergrator for data consistency or Controller for ethical oversight, is discussed with its intent and the specific problem it solves, offering a reusable solution to a recurring design challenge.

Finally, to demonstrate the framework's practical utility, we conduct qualitative case studies on a prominent agent system, ReAct, to diagnose its inherent weaknesses and prescribe targeted improvements using our patterns. 

\section{Design patterns in agentic AI}
The idea of formulating patterns originated as soon as there was a growing trend to adopt FM systems in practice. An article, for instance, suggests seven key patterns (Evals, RAG, Fine-tuning, Caching, Guardrails, Defensive UX) arranged along the lines of enhancing performance versus cutting costs or risk, and getting closer to the data versus the user \cite{yan2023llm-patterns}. In addition, it connects these patterns to the principles of machine learning design such as data flywheel, cascade, and monitoring. The article takes into account software engineering design patterns, includes concrete and specific examples with code, and matches FM patterns with potential problems. Another master's thesis examined current FM applications, including MetaGPT, BabyAGI, and AutoGen as notable examples \cite{ganesh2024exploring}. The six main architectural patterns — Retrieval-Augmented Generation (RAG), In-Context Learning, Ad-hoc, Multi-agent, Usage of Tools, and Chain-of-thought (CoT) prompting — were identified with varying degrees of granularity and presented in the Gang Of Four (GoF)’s format \cite{gamma1995design}, and their applicability to software application development was investigated. In addition, the problems that arise in FM and generative AI also require novel and unique design principles, as demonstrated in the set of six principles for generative AI applications in \cite{Weisz2024DesignPF} and patterns evaluated from practical implementations in \cite{koc2024generative, infoq2025patterns}.

One of the first attempts to categorise design patterns in building AI agents was detailed in a series of blog posts by Andrew Ng \cite{Ng2024Agentic}. These design patterns, namely Reflection, Tool Use, Planning, and Multi-Agent Collaboration, prove to be generalised and simple but effective approaches to enhance the performance and reliability of the system. Following that line, surveys have been conducted based on one or more of these patterns \cite{masterman2024landscape, singh2024enhancing}, and an evolving stack of commonly used tools and subsystems has been observed from AI startups and technology companies' solutions \cite{a16z2023llm, gohel2025ai}. An article takes a step further, designing FM-based agents with security in mind, in order to protect themselves from prompt injection attacks \cite{beurer2025design}. Believing that reliable general-purpose agents are highly improbable, the authors suggest imposing agents with constraints that "explicitly limiting their ability to perform arbitrary tasks", and suggest six design patterns aimed at ensuring a certain degree of isolation between untrusted data and the agent's control flow.

Several notable efforts have proposed comprehensive reference architectures and pattern catalogues, such as the work by Lu et al. \cite{liu2025agent, lu2024towards}. These approaches, often grounded in extensive literature reviews, provide valuable inventories of architectural components and design options.
However, a closer analysis reveals a common characteristic: these architectures are primarily empirically-grounded aggregations of observed functionalities. While practical, this "bottom-up" approach can result in frameworks that lack a unifying theoretical foundation explaining why components interact in a certain way. Furthermore, the "patterns" identified often represent high-level architectural choices (e.g., selecting a plan generator type) rather than reusable, structural solutions to the recurring interaction problems between components, which is the essence of GoF-style patterns.
Our work takes a different, principle-based route. Instead of aggregating existing features, our framework (Section~\ref{sec:concepts}) is derived from the first principles of system theory. This allows us to:
\begin{itemize}
\item Deconstruct an agent into a set of core, interacting subsystems with strong logical coherence.
\item Define granular, interaction-centric design patterns that solve specific collaboration challenges between these subsystems.
\item Emphasise the dynamic flows of information (e.g., context, feedback) that govern the agent's behaviour.
\end{itemize}


We recognise that a classification scheme should include the level of specificity as a key dimension, as being pointed out in \cite{oluyomi2004agent} and \cite{juziuk2014design}. Besides, the relevance of these FM-based agentic AI design patterns to foundational research in design patterns for software, MAS, and AI is also important to be considered. Our classification based on the key literature identified in this section is summarised in Table \ref{tab:dp-lit-comp} below.


\begin{table}[!ht]
    \centering
    \caption{Comparison of literature}
    \label{tab:dp-lit-comp}
    \begin{tabularx}{\textwidth}{|X|c|c|c|c|}
        \hline
        \textbf{Literature} & \textbf{Specificity} & \makecell{\textbf{GoF-}\\\textbf{related}} & \makecell{\textbf{System}\\\textbf{design}} & \textbf{Approach}\\
        \hline
        \cite{ganesh2024exploring} & Specific & $\checkmark$ & $\times$ & Bottom-up\\
        \hline
        \cite{Ng2024Agentic} & General & $\times$ & $\times$ & Top-down \\
        \hline
        \cite{beurer2025design} & Specific & $\checkmark$ & $\times$  & Bottom-up \\
        \hline
        \cite{liu2025agent} & Specific & $\checkmark$ & $\checkmark$  & Bottom-up\\
        \hline
        Ours & Systematic & $\checkmark$ & $\checkmark$ & Integrated \\
        \hline
    \end{tabularx}
\end{table}

The aforementioned publications are vital in shaping our understanding of agentic AI architectures and design patterns. Collectively, they reveal a clear trend towards more structured and reusable solutions. However, this review also highlights a significant gap in the current literature. Most existing approaches do not prioritise a cohesive theoretical foundation, such as system theory, to guide the design and analysis of agent systems. Consequently, the proposed patterns often fall into two categories: either they are high-level strategic concepts (e.g., Ng's four strategies~\cite{Ng2024Agentic}) that lack detailed, implementable structure, or they are specific architectural choices (e.g., CSIRO's catalogue~\cite{liu2025agent}) that, while useful, do not always capture the dynamic, collaborative essence of GoF-style patterns that solve recurring interaction problems.

This gap underscores the need for a framework that is both theoretically grounded and practically applicable through a set of well-defined, structural design patterns in the spirit of the original GoF. Our work aims to fill this gap by proposing:
\begin{itemize}
    \item A system-theoretic agent architecture that explicitly delineates the core functional subsystems and their dynamic interactions.
    \item A collection of agentic design patterns that offer reusable, structural solutions to recurring problems in agent design, emphasising the 'why' and 'how' of inter-subsystem collaboration, not just the 'what' of individual components.
\end{itemize}

\section{Contemporary issues in agentic AI}
\label{sec:challenges}

From the gap analysed above, we review challenges in FM-based agentic AI systems to facilitate the construction of a system design and design patterns. We categorise these problems into five classes with subproblems: World Modelling, Cognitive \& Decision, Execution \& Interaction, Learning \& Governance, and Collaboration Mechanism. These align with human cognitive processes, such as mental modelling, reasoning, action execution, and ethical learning, providing a framework to understand the complexities of agent design. This view is also supported by LeCun's writing "Five Ways to Act Deluded, Stupid, Ineffective, or Evil" which details the five failure modes of agentic AI system based on a human behaviour model \cite{lecun2025fiveways}, and the classification is corroborated by existing literature on this subject. 

\begin{enumerate}
    \item \textbf{World Modelling}: The challenge for FM-based agents is to create an accurate and dynamic representation of their environment. A primary issue is poor \textbf{cognitive data quality}, as models may favour pre-trained knowledge over retrieved information, leading to hallucinations and factual incorrect outputs \cite{xi2025rise, kambhampati2024can}. This is compounded by a lack of \textbf{world model consistency}, where an agent's linguistic competence is "patchy," resulting in logically inconsistent statements \cite{mahowald2024dissociating}. Agents also struggle with \textbf{efficient context retrieval}, not just due to technical limitations on context length \cite{10.5555/3692070.3694435}, but more fundamentally in their inability to reliably integrate retrieved information into their reasoning \cite{du2025rethinking, contextualai2025introducing}. Finally, long-term operation is hindered by challenges in \textbf{state saving and restoring}, where "misaligned experience replay" can cause the propagation of past errors, undermining the agent's performance over time \cite{xiong2025memory}.
    
    \item \textbf{Cognitive \& Decision}: The challenges in cognitive and decision making for agents stem from their probabilistic nature. Regarding \textbf{logical reasoning \& uncertainty}, agents show heuristic aptitude but fail in rigorous and extended logical tasks, while their verbal confidence is an unreliable proxy for actual uncertainty \cite{liu2025logical, han-etal-2024-towards}. This inconsistency is due in part to the lack of a robust internal world model to simulate outcomes \cite{hao-etal-2023-reasoning}. This deficit also affects \textbf{goal-directed behaviour}, where agents struggle to adapt to dynamic environments for long-range goals, as strategies such as task decomposition can be brittle or prone to hallucinations \cite{zheng2025lifelong, zou2025survey, huang2024understanding}. Furthermore, agents are limited by poor \textbf{counterfactual reasoning}, as they often default to pretrained knowledge instead of adapting to contradictory contextual information, restricting their ability to process hypothetical scenarios \cite{yamin2025llms}.
    
    \item \textbf{Execution \& Interaction}: A core challenge is translating plans into reliable real-world actions. Agents often lack \textbf{robustness to environmental changes}, struggling with multimodal perception and amplified hallucinations in chained actions within dynamic settings \cite{tran2025multi}. Although they can be augmented with external tools, their \textbf{effective tool use} is hampered by the difficulty of integrating them into complex workflows, often resulting in non-deterministic "black-box" behaviours that are difficult to debug and control \cite{plaat2025agentic, fournier2025agentic}. This unreliability is exacerbated by inadequate \textbf{error recovery mechanisms}; agents can become trapped in unproductive cycles due to flawed reasoning, and existing reflection methods do not offer guaranteed convergence to a correct solution, especially against adversarial input \cite{huang2024understanding, Kumar2024Certifying}.
    
    \item \textbf{Learning \& Governance}: A key technical hurdle is \textbf{catastrophic forgetting and adaptation to novel situations}, where agents forget previously learnt knowledge when acquiring new data, compromising their ability to adapt without performance degradation \cite{li-etal-2024-revisiting, zheng2025lifelong}. Beyond learning, achieving \textbf{value alignment \& transparency} is a significant challenge, Current alignment methods are costly and can become outdated, while the "black-box" nature of models obscures their reasoning and hinders public trust \cite{padhi-etal-2024-value, calderon-reichart-2025-behalf}. This leads to complex issues in \textbf{ethical choices \& moral development}, as agents lack a human-like understanding of concepts such as intention and can learn unethical behaviours, creating a significant accountability gap for their actions \cite{zou2025survey, reinecke2023puzzle, wang2024ali}.

    \item \textbf{Collaboration Mechanism}: A fundamental obstacle is \textbf{communication and coordination breakdown}, where ambiguous language, asynchronous message sequencing, and security vulnerabilities frequently lead to misinterpretations and failures \cite{Gomez2024HiddenChallenges, tran2025multi, zou2025survey, kong2025survey}. This is compounded by weak \textbf{coordination and joint planning} capabilities; agents often fail to leverage collaboration effectively and exhibit poor joint planning, even in scenarios where cooperation is optimal \cite{ni2025collaborative, agashe-etal-2025-llm}. Finally, navigating complex \textbf{trust and social dynamics} remains a significant hurdle. Building trust is essential for human-agent and inter-agent teams but is often undermined by undesirable emergent social behaviours and the difficulty of managing scenarios involving both cooperation and competition \cite{tran2025multi, ni2025collaborative}.
\end{enumerate}

\section{A system-theoretic agent architecture}
\label{sec:concepts}

To address the systemic challenges outlined previously, we go beyond ad-hoc designs to propose a conceptual framework grounded in system theory. Our approach applies the principle of deconstruction~\cite{bass2003software} to break down an agent into a set of core and extensible subsystems, providing a foundational language to design modular and reliable agents.

\begin{figure}[!ht]
    \centering
    \includegraphics[width=0.9\textwidth]{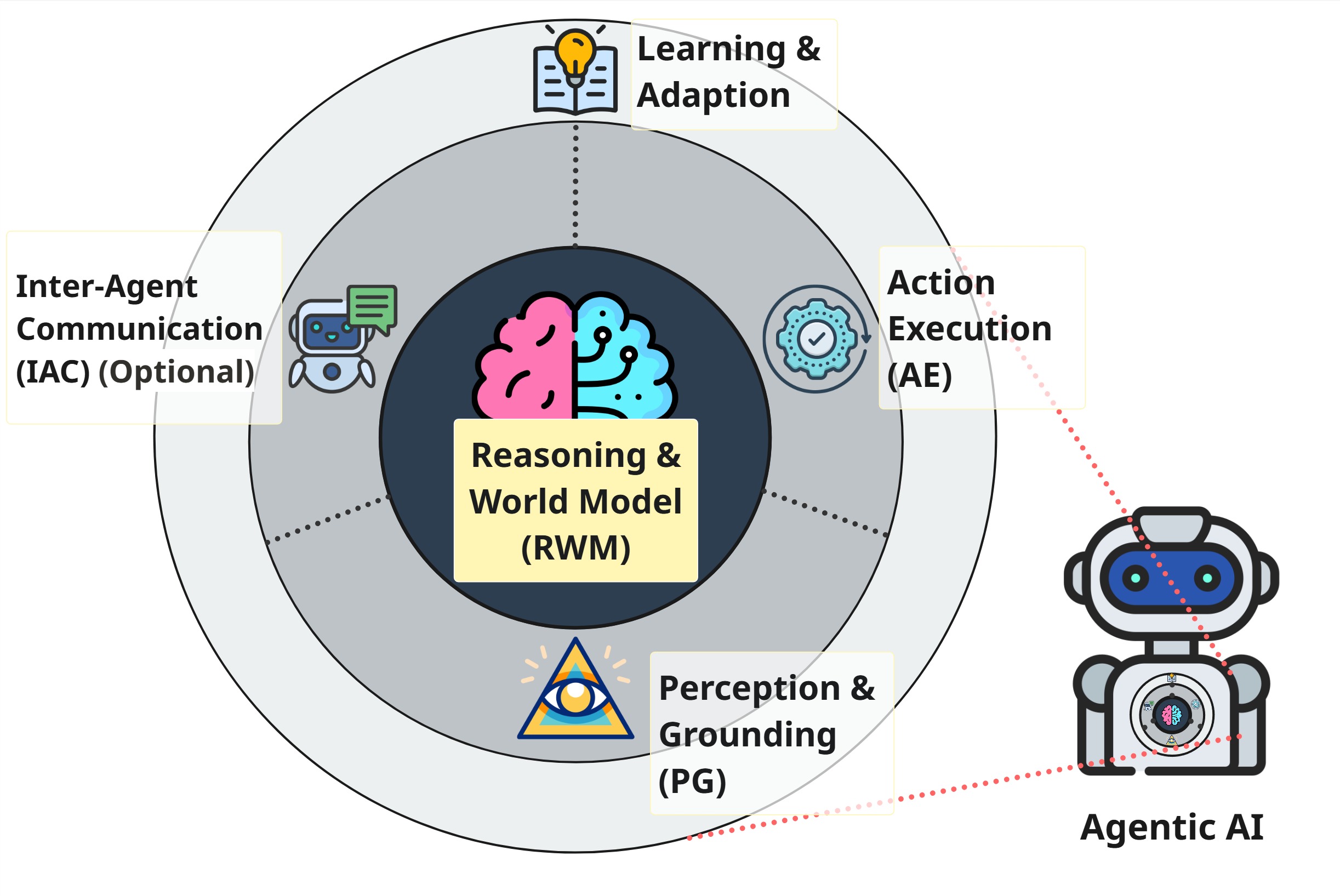}
    \caption{A system-theoretic agent architecture. The model illustrates the internal structure of an agent as nested functional layers, comprising four core subsystems and one extensible subsystem (IAC, highlighted as optional).}
    \label{fig:agent_framework}
\end{figure}

The proposed system-theoretic agent architecture, depicted in Figure~\ref{fig:agent_framework}, visualises this deconstruction. The model conceptualises an agent not as a monolithic entity but as a system of nested functional layers, where each layer represents a different level of abstraction and responsibility. The logic of this layered organisation, which is a direct result of our system-theoretic analysis, is as follows:
\begin{itemize}
    \item \textbf{The cognitive core (innermost layer):} In the centre lies the \texttt{Reasoning \& World Model (RWM)} subsystem. As the agent's decision-making nucleus, it is the most abstract and protected layer, responsible for maintaining the world model and directing all strategic behaviour.
    \item \textbf{The operational interfaces (middle layer):} Surrounding the core is a layer of three subsystems that act as the primary interfaces between the agent's internal reasoning and the external world. This layer includes two fundamental subsystems: the \texttt{Perception \& Grounding (PG)}, which acts as the agent's senses to process and ground raw inputs into percepts, and the \texttt{Action Execution (AE)}, which serves as the agent's effectors to execute actions. For multi-agent capabilities, this layer can be extended with the optional \texttt{Inter-Agent Communication (IAC)} subsystem, the agent's social interface for structured peer-to-peer interaction.
    \item \textbf{The adaptive shell (outermost layer):} Encapsulating the entire system is the \texttt{Learning \& Adaptation (LA)} subsystem. Its position signifies its overarching role: to observe the performance of all inner layers, learn from experience, and drive their continuous improvement through feedback.
\end{itemize}

\begin{figure}[!ht]
    \centering
    \includegraphics[width=0.9\textwidth]{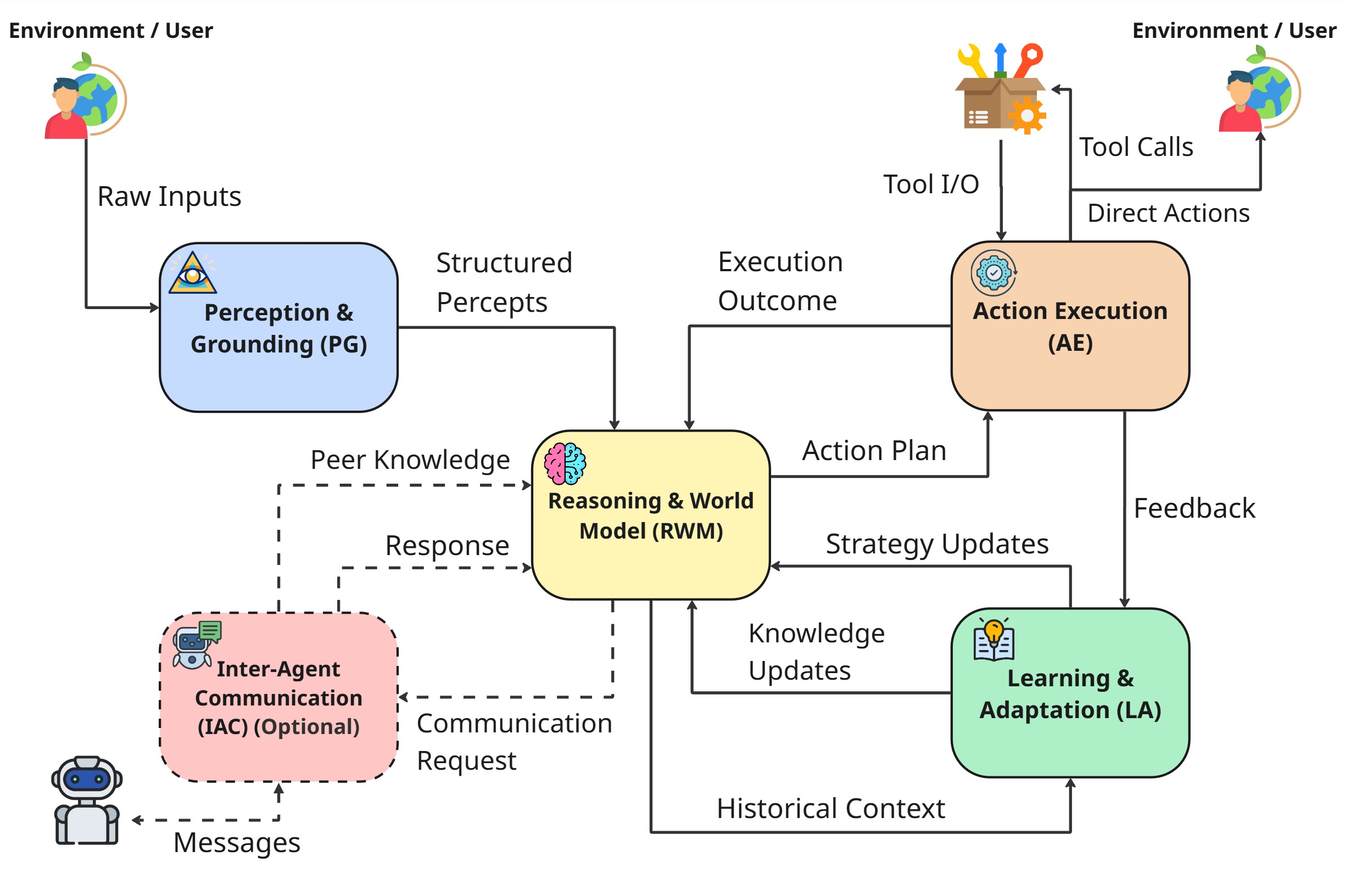} 
    \caption{The agent's cognitive cycle. This diagram illustrates the dynamic interaction flows between the four core subsystems and the optional communication subsystem (\texttt{IAC}, shown with dashed lines).}
    \label{fig:cognitive_cycle}
\end{figure}

Although the system-theoretic agent architecture illustrates the agent's static organisation into functional subsystems, its dynamic operation is best conceptualised as a continuous cognitive cycle. This cycle, a foundational concept in the design of rational agents~\cite{russell2010artificial}, is depicted in Figure~\ref{fig:cognitive_cycle} and details the key information flows that enable intelligent behaviour. The process begins with the \texttt{Perception \& Grounding (PG)} subsystem processing \texttt{Raw Inputs} into \texttt{Structured Percepts}. These percepts are sent to the \texttt{Reasoning \& World Model (RWM)} subsystem, which integrates them to maintain its internal world model. Based on this model, the \texttt{RWM} deliberates and generates either an \texttt{Action Plan} for the \texttt{AE} or a \texttt{Request} for the \texttt{IAC}. The results of these actions generate \texttt{Feedback}, which is processed by the \texttt{LA}. This crucial final step closes the loop: the \texttt{LA} synthesises insights into \texttt{Strategy Updates} and \texttt{Knowledge Updates}, both of which are sent back to the \texttt{RWM} to refine its future reasoning and enrich its world model, allowing true learning and adaptation~\cite{zheng2025lifelong}.

This system-theoretic deconstruction into five core and extensible subsystems provides a stable yet flexible foundation for agent design. It strikes a deliberate trade-off, offering sufficient granularity for analysis while maintaining conceptual clarity. With this architectural blueprint established, we now turn to the specific and reusable solutions for its implementation: the agentic design patterns.

\section{A catalogue of agentic design patterns}
\label{sec:patterns}

The design of agentic AI systems requires a structured and principled approach to address the inherent complexities of autonomy, reliability, and adaptability. Building upon the system-theoretic architecture established in the previous section, we introduce a catalogue of 12 \textit{Agentic Design Patterns} (ADPs).

It is important to note that the concepts underlying many of these patterns are not entirely new; ideas such as reflection, skill acquisition, and tool use have long been explored across various subfields of AI. The primary contribution of this catalogue lies not in the invention of these individual concepts, but in their \textit{systematisation} into a cohesive set of \textit{architectural design patterns} for LLM-based agents. Following our integrated methodology, these patterns are derived both from top-down architectural principles and from a bottom-up analysis of recurring solutions observed in the literature.

Each pattern provides a modular and reusable solution to a recurrent coordination problem among the subsystems of our framework. It establishes a standardised vocabulary and a consistent representational structure (e.g., \textit{Intent}, \textit{Problem}, \textit{Solution}) that describe the involved components, their interactions and practical implications. These patterns are designed to systematically address the identified \textit{Classes of Problems} (Section~\ref{sec:challenges}). Furthermore, they align with Miehling et al.’s~\cite{miehling2025agenticaineedssystems} call for a systems perspective, offering a generative methodology to construct robust and reliable agentic architectures.

\begin{figure}[!ht]
    \centering
    \includegraphics[width=0.75\textwidth]{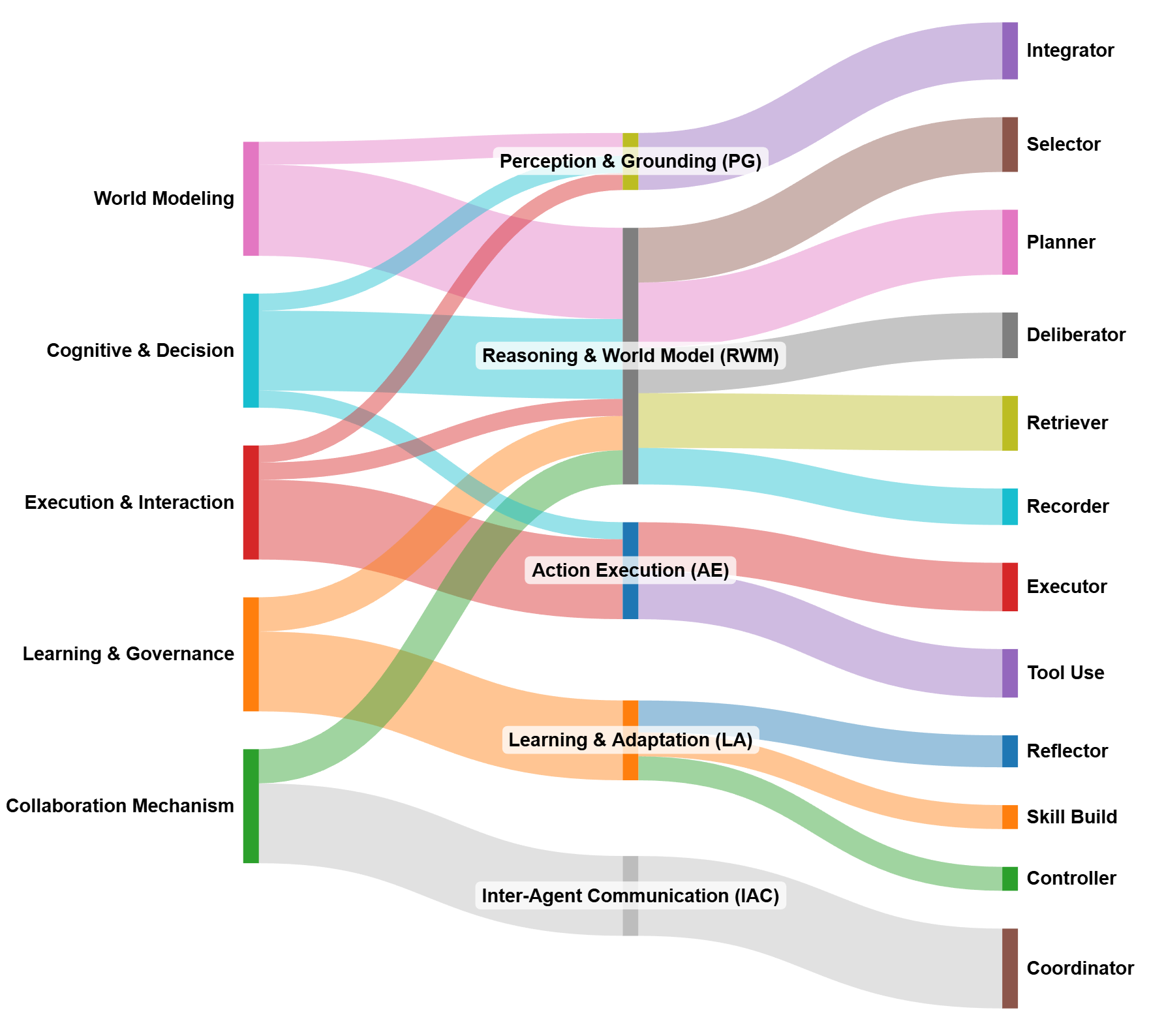} 
    \caption{Conceptual Sankey diagram illustrating the relationships between identified classes of problems, core agent subsystems, and the 12 proposed agentic design patterns. Flow widths indicate qualitative relevance.}
    \label{fig:sankey_diagram}
\end{figure}

To present a holistic view of how these elements interconnect, Figure~\ref{fig:sankey_diagram} illustrates the relationships between the major issues in Section \ref{sec:challenges}, our framework's core subsystems in Section \ref{sec:concepts}, and the ADPs proposed in this section. This Sankey diagram visualises the primary pathways from our identified problem classes to architectural components and finally to specific design solutions. It highlights how World Modelling issues predominantly impact the \texttt{Reasoning \& World Model (RWM)} subsystem, which in turn is addressed by foundational patterns such as \texttt{Integrator} and \texttt{Retriever}.

The complete catalogue is summarised in Table~\ref{tab:pattern_overview}. The patterns are organised into four groups that capture the core aspects of the operation of autonomous agents, reflecting the fundamental components of rational agents as described in the foundational AI literature~\cite{russell2010artificial, georgeff1999belief}. In addition, we briefly describe several representative patterns to illustrate their function and value as follows.

The \texttt{Integrator} pattern addresses cognitive data quality by defining a validation pipeline within the \texttt{PG} subsystem. For decision-making, the \texttt{Selector} pattern provides a solution for adaptive goal-directed behaviour by implementing the Mediator pattern~\cite{gamma1995design} within the \texttt{RWM} to dynamically manage and prioritise the agent's goals. For interaction, the \texttt{Tool Use} pattern ensures effective tool use by acting as a Proxy and Adapter~\cite{gamma1995design} for all external tool calls within the \texttt{AE}. Finally, the \texttt{Controller} pattern addresses value alignment by establishing a continuous monitoring loop, acting as an Observer~\cite{gamma1995design} of the agent's behaviour. A detailed description of all 12 patterns can be found in the full version of our work.

\begin{table}[H]
    \centering
    \caption{Overview of the 12 proposed agentic design patterns (ADPs).}
    \label{tab:pattern_overview}
    \begin{tabularx}{\textwidth}{>{\bfseries}l X >{\raggedright\arraybackslash}X}
    \toprule
    \textbf{Pattern name} & \textbf{Intent} & \textbf{Core problem addressed} \\
    \midrule
    
    \multicolumn{3}{l}{\textit{\textbf{Foundational patterns: building the agent's understanding and state}}} \\
    \textbf{Integrator} & Ensure \texttt{PG} consistency by validating all incoming information. & Cognitive data quality \\
    \textbf{Retriever} & Provide a simplified, context-aware interface to the \texttt{RWM}'s memory. & Inefficient context retrieval \\
    \textbf{Recorder} & Capture and externalise \texttt{RWM} state for later restoration. & State saving \& restoring \\
    \addlinespace
    
    \multicolumn{3}{l}{\textit{\textbf{Cognitive \& decisional patterns: shaping agent thought and action}}} \\
    \textbf{Selector} & Select, prioritise \& adapt primary goal objectives based on dynamic contexts. & Goal-directed behavior (Tactical step selection) \\
    \textbf{Planner} & Decompose high-level goals into manageable, actionable steps. & Goal-directed behavior (Strategic decomposition) \\
    \textbf{Deliberator} & Select the optimal concrete action at each step of the plan. & Goal-directed behavior (Dynamic adaptation) \\
    \addlinespace

    \multicolumn{3}{l}{\textit{\textbf{Execution \& interaction patterns: enabling action and engagement}}} \\
    \textbf{Executor} & Reliably execute the dispatched actions and collect systematic feedback. & Error recovery mechanism \\
    \textbf{Tool Use} & Provide a secure, standardised interface for all external tool invocations. & Effective tool use \\
    \textbf{Coordinator} & Manage and facilitate structured multi-agent communication. & Communication and coordination breakdown \\
    \addlinespace
    
    \multicolumn{3}{l}{\textit{\textbf{Adaptive \& learning patterns: enabling improvement and evolution}}} \\
    \textbf{Reflector} & Analyse outcomes to infer causality and generate actionable insights. & Adaptation (Causal learning) \\
    \textbf{Skill Build} & Discover and refine reusable procedural skills from experience. & Adaptation (Procedural learning) \\
    \textbf{Controller} & Continuously monitor and align agent behaviour with ethical principles. & Value alignment \& transparency \\
    
    \bottomrule
    \end{tabularx}
\end{table}

\section{Application: a qualitative analysis of an existing system}

This section uses a bottom-up approach to validate our framework, qualitatively analysing the ReAct system to demonstrate enhancements from our system-theoretic architecture through a three-step methodology.
\begin{enumerate}
    \item \textbf{Deconstruct:} We deconstruct the architecture to map its functionalities to the five core subsystems of the framework.
    \item \textbf{Diagnose:} We then diagnose architectural weaknesses by analysing which of our five problem classes manifest most prominently in the system.
    \item \textbf{Prescribe:} We propose specific agentic design patterns (ADPs) as solutions to these problems.
\end{enumerate}

\paragraph{Deconstruct:} In the ReAct paradigm, the functionalities of our subsystems are implicitly and monolithically implemented within the central LLM and its interaction loop.
\begin{itemize}
    \item The \textit{Reasoning \& World Model (RWM)} is similar to the LLM's `Thought` generation process. Its world model is an implicit and transient state held within the LLM's limited context window.
    \item The \textit{Perception \& Grounding (PG)} is rudimentary; the agent perceives the world solely through unstructured `Observations` from the environment.
    \item The \textit{Action Execution (AE)} is the `Act` step, where the LLM's generated action is passed to an external environment or tool.
    \item The \textit{Learning \& Adaptation (LA)} and \textit{Inter-Agent Communication (IAC)} are absent. ReAct has no mechanism for long-term learning and is designed as a single-agent framework.
\end{itemize}

\paragraph{Diagnose:} The original ReAct framework exhibits systemic fragilities. Its monolithic design leads to significant world-modelling challenges, as it lacks mechanisms for validating observations or managing context efficiently. The unstructured `Thought` process results in suboptimal planning, hindering goal-directed behaviour. Furthermore, the framework lacks robust error recovery mechanisms for tool use and a dedicated process for adaptation, which prevents it from learning from failures.

\begin{figure}[H]
    \centering
    \includegraphics[width=0.8\textwidth]{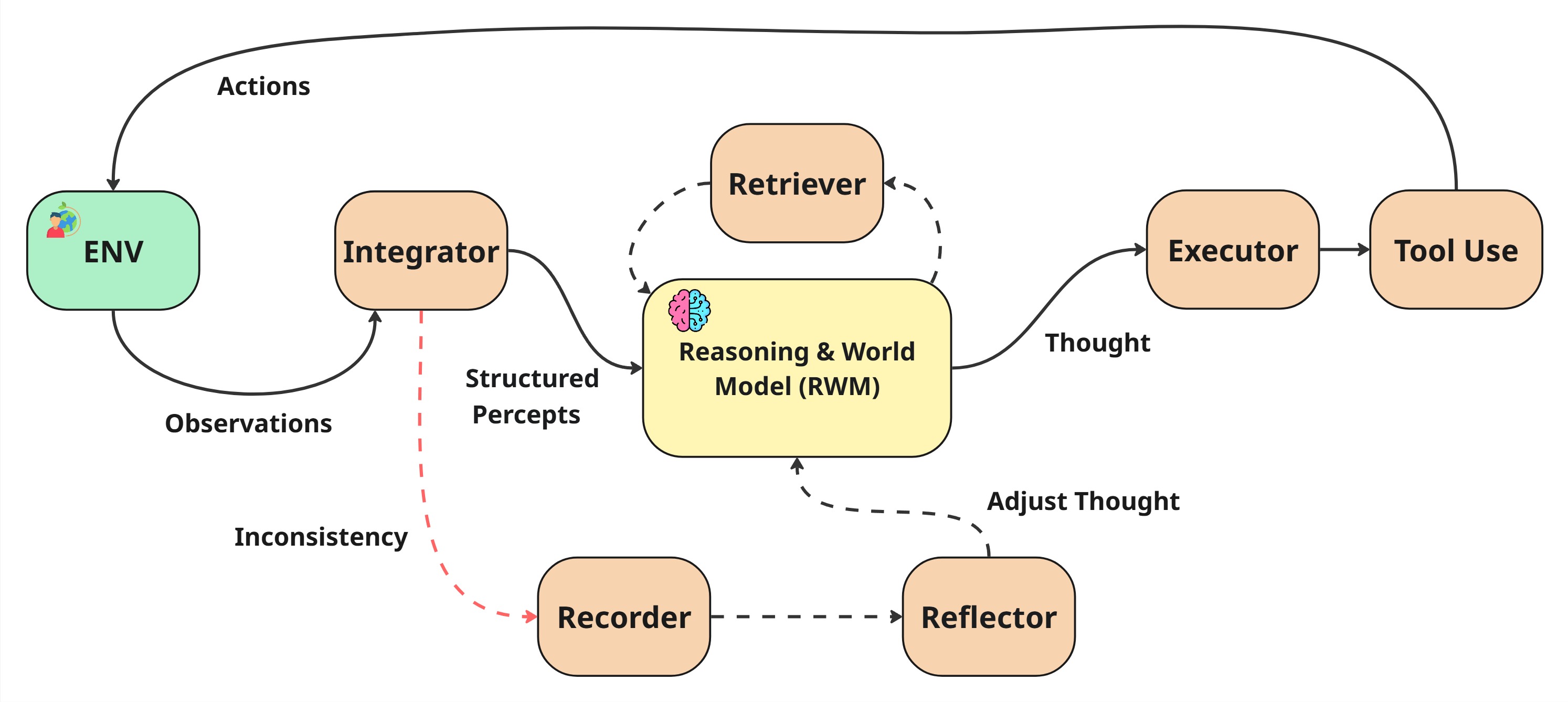}
    \caption{A conceptual diagram showing how ReAct can be enhanced by integrating our proposed agentic design patterns.}
    \label{fig:react_improved_flow}
\end{figure}

\paragraph{Prescribe:} We propose enhancing the ReAct loop by integrating specific ADPs, transforming its simple cycle into a structured workflow as illustrated in Figure~\ref{fig:react_improved_flow}. To address World Modelling challenges, the Integrator pattern first validates the incoming observations. In case of a critical \texttt{Inconsistency}, it can trigger the Recorder to save the problematic state and the Reflector to initiate a learning cycle. For valid data, the Retriever and Recorder patterns provide robust mechanisms for context retrieval and state management within the core \texttt{RWM}. The \texttt{RWM}'s \texttt{Thought} is then passed to the Executor and Tool Use patterns for reliable execution. Finally, the feedback from this execution is also processed by the Reflector, allowing the agent to perform causal analysis of failures and adjust future strategies, creating an adaptive and resilient agent.

\section{Limitations and future work}
We acknowledge several limitations that also highlight promising directions for future research. Our framework is primarily conceptual; a critical next step is to conduct quantitative benchmarking to empirically measure the performance improvements (e.g., reliability, efficiency) offered by our patterns against baselines. Secondly, the implementation of sophisticated patterns such as Reflector and Controller introduces architectural complexity and potential computational overhead, whose trade-offs require further investigation. Finally, while our work promotes reliable agent design, it does not fully address the broader societal impacts of large-scale autonomous systems, such as accountability and emergent behaviours, which remain open critical problems.

\section{Conclusion}
The rapid development of agentic AI has largely relied on ad-hoc methods, resulting in systems that are powerful but often brittle and unreliable. This paper addresses this fundamental issue by introducing a principled engineering discipline grounded in system theory. We proposed a novel framework that deconstructs an agent into five core subsystems and presented a catalogue of 12 agentic design patterns that offer structural solutions to recurring problems. The practical utility of this approach was demonstrated through our qualitative analysis of the ReAct framework, where we diagnosed its systemic weaknesses and prescribed specific patterns to enhance robustness and adaptability. By providing a shared vocabulary and a structured methodology, this work aims to shift the development of agentic systems from informal experimentation to a principled engineering practice, paving the way for more modular, reliable and trustworthy autonomous agents.

\begin{ack}
This publication has emanated from research supported in part by grants from Research Ireland under Grant 12-RC-2289-P2 and 18/CRT/6223 which is co-funded under the European Regional Development Fund. For the purpose of Open Access, the author has applied a CC BY-NC 4.0 public copyright licence to any Author Accepted Manuscript version arising from this submission.
\end{ack}

\bibliographystyle{unsrtnat}
\bibliography{sn-bibliography}

@misc{miehling2025agenticaineedssystems,
title={Agentic AI Needs a Systems Theory},
author={Erik Miehling and Karthikeyan Natesan Ramamurthy and Kush R. Varshney and Matthew Riemer and Djallel Bouneffouf and John T. Richards and Amit Dhurandhar and Elizabeth M. Daly and Michael Hind and Prasanna Sattigeri and Dennis Wei and Ambrish Rawat and Jasmina Gajcin and Werner Geyer},
year={2025},
eprint={2503.00237},
archivePrefix={arXiv},
primaryClass={cs.AI},
url={https://arxiv.org/abs/2503.00237},
}

@book{russell2010artificial,
  title     = {Artificial Intelligence: A Modern Approach},
  author    = {Russell, Stuart J. and Norvig, Peter},
  year      = {2010},
  edition   = {3},
  publisher = {Pearson Education},
  address   = {Upper Saddle River, NJ, USA},
  isbn      = {978-0136042594}
}

@misc{Ng2024Agentic,
  author   = {Andrew Ng},
  title    = {Agentic Design Patterns Part~1: Four AI Agent Strategies That Improve GPT-4 and GPT-3.5 Performance},
  year     = {2024},
  month    = {mar},
  url      = {https://www.deeplearning.ai/the-batch/how-agents-can-improve-llm-performance/},
  urldate  = {2025-05-21},
  organization = {DeepLearning.AI},
  note     = {The Batch (blog)}
}

@misc{lecun2025fiveways,
  author       = {Yann LeCun},
  title        = {Five Ways to Act Deluded, Stupid, Ineffective, or Evil},
  year         = {2025},
  month        = {may},  
  day          = {10},   
  url          = {https://www.linkedin.com/posts/yann-lecun_five-ways-to-act-deluded-stupid-ineffective-activity-7327058733967052800-TsUK/},
  urldate  = {2025-05-22},
  organization = {LinkedIn}
}

@article{yan2023llm-patterns,
  title   = {Patterns for Building LLM-based Systems \& Products},
  author  = {Yan, Ziyou},
  journal = {eugeneyan.com},
  year    = {2023},
  month   = {Jul},
  url     = {https://eugeneyan.com/writing/llm-patterns/}
}

@mastersthesis{ganesh2024exploring,
  title={Exploring Patterns in LLM Integration-A study on architectural considerations and design patterns in LLM dependent applications},
  author={Ganesh, Sundarakrishnan and Sahlqvist, Robert},
  school={CHALMERS UNIVERSITY OF TECHNOLOGY, Gothenburg, Sweden}, 
  year={2024}
}

@article{masterman2024landscape,
  title={The landscape of emerging ai agent architectures for reasoning, planning, and tool calling: A survey},
  author={Masterman, Tula and Besen, Sandi and Sawtell, Mason and Chao, Alex},
  journal={arXiv preprint arXiv:2404.11584},
  year={2024}
}

@inproceedings{singh2024enhancing,
  title={Enhancing ai systems with agentic workflows patterns in large language model},
  author={Singh, Aditi and Ehtesham, Abul and Kumar, Saket and Khoei, Tala Talaei},
  booktitle={2024 IEEE World AI IoT Congress (AIIoT)},
  pages={527--532},
  year={2024},
  organization={IEEE}
}

@book{gamma1995design,
    author = {Gamma, Erich and Helm, Richard and Johnson, Ralph and Vlissides, John},
    title = {Design patterns: elements of reusable object-oriented software},
    year = {1995},
    isbn = {0201633612},
    publisher = {Addison-Wesley Longman Publishing Co., Inc.},
    address = {USA}
}

@misc{a16z2023llm,
  author = {{Andreessen Horowitz}},
  title = {Emerging Architectures for LLM Applications},
  year = {2023},
  url = {https://a16z.com/emerging-architectures-for-llm-applications/},
  urldate = {2025-06-05}
}

@misc{gohel2025ai,
  author = {Gohel, Rakesh},
  title = {Don't waste every day reinventing your AI Agent Architecture},
  year = {2025},
  url = {https://www.linkedin.com/posts/rakeshgohel01_dont-waste-every-day-reinventing-your-ai-activity-7331296814974808065-co8M/},
  urldate = {2025-06-05}
}

@inproceedings{georgeff1999belief,
  title={The belief-desire-intention model of agency},
  author={Georgeff, Michael and Pell, Barney and Pollack, Martha and Tambe, Milind and Wooldridge, Michael},
  booktitle={Intelligent Agents V: Agents Theories, Architectures, and Languages: 5th International Workshop, ATAL’98 Paris, France, July 4--7, 1998 Proceedings 5},
  pages={1--10},
  year={1999},
  organization={Springer}
}

@article{Weisz2024DesignPF,
  title={Design Principles for Generative AI Applications},
  author={Justin D. Weisz and Jessica He and Michael Muller and Gabriela Hoefer and Rachel Miles and Werner Geyer},
  journal={Proceedings of the 2024 CHI Conference on Human Factors in Computing Systems},
  year={2024},
  url={https://api.semanticscholar.org/CorpusID:267301068}
}

@article{tran2025multi,
  title={Multi-Agent Collaboration Mechanisms: A Survey of LLMs},
  author={Tran, Khanh-Tung and Dao, Dung and Nguyen, Minh-Duong and Pham, Quoc-Viet and O'Sullivan, Barry and Nguyen, Hoang D},
  journal={arXiv preprint arXiv:2501.06322},
  year={2025}
}

@article{koc2024generative,
  author = {Koc, Vincent},
  title = {Generative AI Design Patterns: A Comprehensive Guide},
  journal = {Towards Data Science},
  year = {2024},
  month = {2},
  day = {13},
  url = {https://medium.com/data-science/generative-ai-design-patterns-a-comprehensive-guide-41425a40d7d0}
}

@article{infoq2025patterns,
  author = {Rahul Suresh},
  title = {Beyond the Gang of Four: Practical Design Patterns for Modern AI Systems},
  journal = {InfoQ},
  year = {2025},
  url = {https://www.infoq.com/articles/practical-design-patterns-modern-ai-systems/}
}

@inproceedings{oluyomi2004agent,
  title={An agent design pattern classification scheme: Capturing the notions of agency in agent design patterns},
  author={Oluyomi, Ayodele and Karunasekera, Shanika and Sterling, Leon},
  booktitle={11th Asia-Pacific Software Engineering Conference},
  pages={456--463},
  year={2004},
  organization={IEEE}
}

@article{juziuk2014design,
  title={Design patterns for multi-agent systems: A systematic literature review},
  author={Juziuk, Joanna and Weyns, Danny and Holvoet, Tom},
  journal={Agent-oriented software engineering: reflections on architectures, methodologies, languages, and frameworks},
  pages={79--99},
  year={2014},
  publisher={Springer}
}

@article{liu2025agent,
  title={Agent design pattern catalogue: A collection of architectural patterns for foundation model based agents},
  author={Liu, Yue and Lo, Sin Kit and Lu, Qinghua and Zhu, Liming and Zhao, Dehai and Xu, Xiwei and Harrer, Stefan and Whittle, Jon},
  journal={Journal of Systems and Software},
  volume={220},
  pages={112278},
  year={2025},
  publisher={Elsevier}
}

@article{beurer2025design,
  title={Design Patterns for Securing LLM Agents against Prompt Injections},
  author={Beurer-Kellner, Luca and Cre{\c{t}}u, Beat Buesser Ana-Maria and Debenedetti, Edoardo and Dobos, Daniel and Fabian, Daniel and Fischer, Marc and Froelicher, David and Grosse, Kathrin and Naeff, Daniel and Ozoani, Ezinwanne and others},
  journal={arXiv preprint arXiv:2506.08837},
  year={2025}
}

@inproceedings{lu2024towards,
  title={Towards responsible generative ai: A reference architecture for designing foundation model based agents},
  author={Lu, Qinghua and Zhu, Liming and Xu, Xiwei and Xing, Zhenchang and Harrer, Stefan and Whittle, Jon},
  booktitle={2024 IEEE 21st International Conference on Software Architecture Companion (ICSA-C)},
  pages={119--126},
  year={2024},
  organization={IEEE}
}

@article{xi2025rise,
  title={The rise and potential of large language model based agents: A survey},
  author={Xi, Zhiheng and Chen, Wenxiang and Guo, Xin and He, Wei and Ding, Yiwen and Hong, Boyang and Zhang, Ming and Wang, Junzhe and Jin, Senjie and Zhou, Enyu and others},
  journal={Science China Information Sciences},
  volume={68},
  number={2},
  pages={121101},
  year={2025},
  publisher={Springer}
}

@article{plaat2025agentic,
  title={Agentic large language models, a survey},
  author={Plaat, Aske and van Duijn, Max and van Stein, Niki and Preuss, Mike and van der Putten, Peter and Batenburg, Kees Joost},
  journal={arXiv preprint arXiv:2503.23037},
  year={2025}
}

@article{kambhampati2024can,
  title={Can large language models reason and plan?},
  author={Kambhampati, Subbarao},
  journal={Annals of the New York Academy of Sciences},
  volume={1534},
  number={1},
  pages={15--18},
  year={2024},
  publisher={Wiley Online Library}
}

@article{mahowald2024dissociating,
  title={Dissociating language and thought in large language models},
  author={Mahowald, Kyle and Ivanova, Anna A and Blank, Idan A and Kanwisher, Nancy and Tenenbaum, Joshua B and Fedorenko, Evelina},
  journal={Trends in cognitive sciences},
  year={2024},
  publisher={Elsevier}
}

@article{du2025rethinking,
  title={Rethinking memory in ai: Taxonomy, operations, topics, and future directions},
  author={Du, Yiming and Huang, Wenyu and Zheng, Danna and Wang, Zhaowei and Montella, Sebastien and Lapata, Mirella and Wong, Kam-Fai and Pan, Jeff Z},
  journal={arXiv preprint arXiv:2505.00675},
  year={2025}
}

@misc{contextualai2025introducing,
  author       = {Contextual AI Team},
  title        = {Introducing the most grounded language model in the world},
  year         = {2025},
  howpublished = {\url{https://contextual.ai/blog/introducing-grounded-language-model/}},
  note         = {Accessed: 2025-06-30}
}

@misc{10.5555/3692070.3694435,
      title={When Linear Attention Meets Autoregressive Decoding: Towards More Effective and Efficient Linearized Large Language Models}, 
      author={Haoran You and Yichao Fu and Zheng Wang and Amir Yazdanbakhsh and Yingyan Celine Lin},
      year={2024},
      eprint={2406.07368},
      archivePrefix={arXiv},
      primaryClass={cs.CL},
      url={https://arxiv.org/abs/2406.07368}, 
}

@article{xiong2025memory,
  title={How Memory Management Impacts LLM Agents: An Empirical Study of Experience-Following Behavior},
  author={Xiong, Zidi and Lin, Yuping and Xie, Wenya and He, Pengfei and Tang, Jiliang and Lakkaraju, Himabindu and Xiang, Zhen},
  journal={arXiv preprint arXiv:2505.16067},
  year={2025}
}

@article{liu2025logical,
  title={Logical Reasoning in Large Language Models: A Survey},
  author={Liu, Hanmeng and Fu, Zhizhang and Ding, Mengru and Ning, Ruoxi and Zhang, Chaoli and Liu, Xiaozhang and Zhang, Yue},
  journal={arXiv preprint arXiv:2502.09100},
  year={2025}
}

@inproceedings{hao-etal-2023-reasoning,
    title = "Reasoning with Language Model is Planning with World Model",
    author = "Hao, Shibo  and
      Gu, Yi  and
      Ma, Haodi  and
      Hong, Joshua  and
      Wang, Zhen  and
      Wang, Daisy  and
      Hu, Zhiting",
    editor = "Bouamor, Houda  and
      Pino, Juan  and
      Bali, Kalika",
    booktitle = "Proceedings of the 2023 Conference on Empirical Methods in Natural Language Processing",
    month = dec,
    year = "2023",
    address = "Singapore",
    publisher = "Association for Computational Linguistics",
    url = "https://aclanthology.org/2023.emnlp-main.507/",
    doi = "10.18653/v1/2023.emnlp-main.507",
    pages = "8154--8173"
}

@inproceedings{han-etal-2024-towards,
    title = "Towards Uncertainty-Aware Language Agent",
    author = "Han, Jiuzhou  and
      Buntine, Wray  and
      Shareghi, Ehsan",
    editor = "Ku, Lun-Wei  and
      Martins, Andre  and
      Srikumar, Vivek",
    booktitle = "Findings of the Association for Computational Linguistics: ACL 2024",
    month = aug,
    year = "2024",
    address = "Bangkok, Thailand",
    publisher = "Association for Computational Linguistics",
    url = "https://aclanthology.org/2024.findings-acl.398/",
    doi = "10.18653/v1/2024.findings-acl.398",
    pages = "6662--6685"
}

@article{zheng2025lifelong,
  title={Lifelong Learning of Large Language Model based Agents: A Roadmap},
  author={Zheng, Junhao and Shi, Chengming and Cai, Xidi and Li, Qiuke and Zhang, Duzhen and Li, Chenxing and Yu, Dong and Ma, Qianli},
  journal={arXiv preprint arXiv:2501.07278},
  year={2025}
}

@article{zou2025survey,
  title={A survey on large language model based human-agent systems},
  author={Zou, Henry Peng and Huang, Wei-Chieh and Wu, Yaozu and Chen, Yankai and Miao, Chunyu and Nguyen, Hoang and Zhou, Yue and Zhang, Weizhi and Fang, Liancheng and He, Langzhou and others},
  journal={arXiv preprint arXiv:2505.00753},
  year={2025}
}

@article{huang2024understanding,
  title={Understanding the planning of LLM agents: A survey},
  author={Huang, Xu and Liu, Weiwen and Chen, Xiaolong and Wang, Xingmei and Wang, Hao and Lian, Defu and Wang, Yasheng and Tang, Ruiming and Chen, Enhong},
  journal={arXiv preprint arXiv:2402.02716},
  year={2024}
}

@misc{ni2025collaborative,
  author       = {Ni, Ansong and Desai, Ruta and Li, Yang and Lei, Xinjie and Wang, Dong and Raghavendra, Ramya and Ghosh, Gargi and Li, Daniel and Celikyilmaz, Asli},
  title        = {Collaborative Reasoner: Self-improving Social Agents with Synthetic Conversations},
  year         = {2025},
  howpublished = {\url{https://ai.meta.com/research/publications/collaborative-reasoner-self-improving-social-agents-with-synthetic-conversations/}},
  note         = {Accessed: 2025-06-30}
}

@article{yamin2025llms,
  title={LLMs Struggle to Perform Counterfactual Reasoning with Parametric Knowledge},
  author={Yamin, Khurram and Ghosal, Gaurav and Wilder, Bryan},
  journal={arXiv preprint arXiv:2506.15732},
  year={2025}
}

@article{fournier2025agentic,
  title={Agentic AI Process Observability: Discovering Behavioral Variability},
  author={Fournier, Fabiana and Limonad, Lior and David, Yuval},
  journal={arXiv preprint arXiv:2505.20127},
  year={2025}
}

@article{Kumar2024Certifying,
  author = {Aounon Kumar and Chirag Agarwal and Suraj Srinivas and Aaron Jiaxun Li and Soheil Feizi and Himabindu Lakkaraju},
  title = {Certifying {LLM} Safety against Adversarial Prompting},
  journal = {Conference on Language Modeling (COLM)},
  year = {2024},
  url = {https://openreview.net/pdf?id=9Ik05cycLq}
}

@inproceedings{li-etal-2024-revisiting,
    title = "Revisiting Catastrophic Forgetting in Large Language Model Tuning",
    author = "Li, Hongyu  and
      Ding, Liang  and
      Fang, Meng  and
      Tao, Dacheng",
    editor = "Al-Onaizan, Yaser  and
      Bansal, Mohit  and
      Chen, Yun-Nung",
    booktitle = "Findings of the Association for Computational Linguistics: EMNLP 2024",
    month = nov,
    year = "2024",
    address = "Miami, Florida, USA",
    publisher = "Association for Computational Linguistics",
    url = "https://aclanthology.org/2024.findings-emnlp.249/",
    doi = "10.18653/v1/2024.findings-emnlp.249",
    pages = "4297--4308"
}

@inproceedings{padhi-etal-2024-value,
    title = "Value Alignment from Unstructured Text",
    author = "Padhi, Inkit  and
      Natesan Ramamurthy, Karthikeyan  and
      Sattigeri, Prasanna  and
      Nagireddy, Manish  and
      Dognin, Pierre  and
      Varshney, Kush R.",
    editor = "Dernoncourt, Franck  and
      Preo{\c{t}}iuc-Pietro, Daniel  and
      Shimorina, Anastasia",
    booktitle = "Proceedings of the 2024 Conference on Empirical Methods in Natural Language Processing: Industry Track",
    month = nov,
    year = "2024",
    address = "Miami, Florida, US",
    publisher = "Association for Computational Linguistics",
    url = "https://aclanthology.org/2024.emnlp-industry.81/",
    doi = "10.18653/v1/2024.emnlp-industry.81",
    pages = "1083--1095"
}

@inproceedings{calderon-reichart-2025-behalf,
    title = "On Behalf of the Stakeholders: Trends in {NLP} Model Interpretability in the Era of {LLM}s",
    author = "Calderon, Nitay  and
      Reichart, Roi",
    editor = "Chiruzzo, Luis  and
      Ritter, Alan  and
      Wang, Lu",
    booktitle = "Proceedings of the 2025 Conference of the Nations of the Americas Chapter of the Association for Computational Linguistics: Human Language Technologies (Volume 1: Long Papers)",
    month = apr,
    year = "2025",
    address = "Albuquerque, New Mexico",
    publisher = "Association for Computational Linguistics",
    url = "https://aclanthology.org/2025.naacl-long.29/",
    doi = "10.18653/v1/2025.naacl-long.29",
    pages = "656--693",
    ISBN = "979-8-89176-189-6"
}

@article{reinecke2023puzzle,
  title={The puzzle of evaluating moral cognition in artificial agents},
  author={Reinecke, Madeline G and Mao, Yiran and Kunesch, Markus and Du{\'e}{\~n}ez-Guzm{\'a}n, Edgar A and Haas, Julia and Leibo, Joel Z},
  journal={Cognitive Science},
  volume={47},
  number={8},
  pages={e13315},
  year={2023},
  publisher={Wiley Online Library}
}

@article{wang2024ali,
  title={Ali-agent: Assessing llms' alignment with human values via agent-based evaluation},
  author={Wang, Han and Zhang, An and Duy Tai, Nguyen and Sun, Jun and Chua, Tat-Seng and others},
  journal={Advances in Neural Information Processing Systems},
  volume={37},
  pages={99040--99088},
  year={2024}
}

@article{Gomez2024HiddenChallenges,
  author  = {Kye Gomez},
  title   = {The Hidden Challenges of Multi-LLM Agent Collaboration},
  journal = {Medium},
  year    = {2024},
  month   = {sep},
  day     = {30},
  url     = {https://medium.com/@kyeg/the-hidden-challenges-of-multi-llm-agent-collaboration-59c83f347503},
  urldate = {2025-07-01}
}

@article{kong2025survey,
  title={A Survey of LLM-Driven AI Agent Communication: Protocols, Security Risks, and Defense Countermeasures},
  author={Kong, Dezhang and Lin, Shi and Xu, Zhenhua and Wang, Zhebo and Li, Minghao and Li, Yufeng and Zhang, Yilun and Sha, Zeyang and Li, Yuyuan and Lin, Changting and others},
  journal={arXiv preprint arXiv:2506.19676},
  year={2025}
}

@inproceedings{agashe-etal-2025-llm,
    title = "{LLM}-Coordination: Evaluating and Analyzing Multi-agent Coordination Abilities in Large Language Models",
    author = "Agashe, Saaket  and
      Fan, Yue  and
      Reyna, Anthony  and
      Wang, Xin Eric",
    editor = "Chiruzzo, Luis  and
      Ritter, Alan  and
      Wang, Lu",
    booktitle = "Findings of the Association for Computational Linguistics: NAACL 2025",
    month = apr,
    year = "2025",
    address = "Albuquerque, New Mexico",
    publisher = "Association for Computational Linguistics",
    url = "https://aclanthology.org/2025.findings-naacl.448/",
    doi = "10.18653/v1/2025.findings-naacl.448",
    pages = "8038--8057",
    ISBN = "979-8-89176-195-7"
}

@book{bass2003software,
author = {Bass, Len and Clements, Paul and Kazman, Rick},
title = {Software Architecture in Practice},
year = {2003},
isbn = {0321154959},
publisher = {Addison-Wesley Longman Publishing Co., Inc.},
address = {USA},
edition = {2}
}

@article{naveed2025comprehensive,
  title={A comprehensive overview of large language models},
  author={Naveed, Humza and Khan, Asad Ullah and Qiu, Shi and Saqib, Muhammad and Anwar, Saeed and Usman, Muhammad and Akhtar, Naveed and Barnes, Nick and Mian, Ajmal},
  journal={ACM Transactions on Intelligent Systems and Technology},
  volume={16},
  number={5},
  pages={1--72},
  year={2025},
  publisher={ACM New York, NY}
}

@article{kaddour2023challenges,
  title={Challenges and applications of large language models},
  author={Kaddour, Jean and Harris, Joshua and Mozes, Maximilian and Bradley, Herbie and Raileanu, Roberta and McHardy, Robert},
  journal={arXiv preprint arXiv:2307.10169},
  year={2023}
}

@article{acharya2025agentic,
  title={Agentic ai: Autonomous intelligence for complex goals--a comprehensive survey},
  author={Acharya, Deepak Bhaskar and Kuppan, Karthigeyan and Divya, B},
  journal={IEEE Access},
  year={2025},
  publisher={IEEE}
}

\end{document}